%% file: main.tex
\documentclass[10pt,twocolumn]{article}

\usepackage[margin=0.78in]{geometry}
\usepackage{amsmath,amssymb}
\usepackage{booktabs}
\usepackage{etoolbox}
\usepackage{microtype}
\usepackage{tikz}
\usepackage[numbers,sort&compress]{natbib}
\PassOptionsToPackage{hyphens}{url}
\usepackage[hidelinks]{hyperref}

\title{TRACE-C: Rank-Calibrated Relational Anomaly Detection\\
       for Multi-Stream Operational Telemetry}
\author{Matthew Faucher\\Independent researcher}
\date{19 Aug 2026}

\begin{document}
\maketitle

\begin{abstract}
\input{sections/abstract}
\end{abstract}

\input{sections/introduction}
\input{sections/related-work}
\input{sections/method}
\input{sections/data-protocol}
\input{sections/results}
\input{sections/limitations}
\input{sections/conclusion}

\section*{Code and data availability}
The implementation, committed machine-readable reports, table generator, and
reproduction instructions accompany this manuscript. Retained telemetry inputs
are covered by checksums and report-generator provenance. Data are provided by
the National Energy System Operator under the NESO Open Data Licence
\citep{nesoOpenLicence}. Required attribution: ``Supported by National Energy
SO Open Data.'' Preprint:
\url{https://doi.org/10.5281/zenodo.22012123}.
Code and reproduction instructions:
\url{https://github.com/mars-arch/trace-c}. Evidence package:
\url{https://huggingface.co/datasets/mars2titan/trace-c-neso-benchmark},
DOI \url{https://doi.org/10.57967/hf/10063}.

\bibliographystyle{plainnat}
\bibliography{references}
\end{document}

%% file: sections/abstract.tex
Operational telemetry can be jointly anomalous while every individual stream
stays inside its familiar range. TRACE-C is an auditable strictly-prior
rank-calibrated detector for aligned multi-stream telemetry: same-regime
rolling median/MAD residuals feed three window channels---a maximum normalized
local sum, a Gaussian copula-form dependence contrast on robust-z residuals,
and a worst standardized AR(1) innovation---whose channel ranks are
Fisher-aggregated and ranked against earlier aggregates.

We evaluate six Great Britain grid streams with a January--April 2019 fit,
July--December 2019 development evidence, and a 2020 hold-out frozen before
inspection. TRACE-C ranks Storm Atiyah first among 2019 test windows, but a
disclosed channel ablation attributes that rank to the local channel, not the
copula-form channel: copula-only ranks Atiyah 59th. The short 9 August
frequency event is ranked far
lower by the fused detector (143) than by the temporal channel alone (40), and
reconstruction baselines rank it first. In 2020 no window is selected, which is
consistent with record-rule saturation rather than an uneventful year; the
highest-ranked frozen window was later interpreted as Storm Ellen.

Three interpretive limits carry throughout. The resulting $p$-values are
selection quantities, not event probabilities. The copula-form channel is not a
literal copula density: the method applies no probability-integral or
normal-score transform. Empirical rank counts are diagnostics, not coverage or
false-discovery proofs. Every table and figure in this paper is generated from
committed machine-readable reports.

%% file: sections/introduction.tex
\section{Introduction}

Operational monitoring is intrinsically multivariate. Demand, embedded
generation, pumping, and frequency can each remain within a historically
familiar range while their joint configuration changes. Conversely, seasonal
or regime drift can make many conventional anomaly scores large without a
discrete operational event. The practical question is therefore not only
whether a value is extreme, but whether a window is unusual relative to the
information that would actually have been available when it arrived.

That question is difficult for three reasons. First, telemetry is serially
dependent and nonstationary, whereas simple rank arguments are sharpest under
exchangeability \citep{shafer2008tutorial}. Second, coarse summaries may erase
short physical transients even when they make longer multi-stream episodes
easy to rank. Third, an operational detector includes a selection policy as
well as a score. A nominal multiple-testing procedure, a fallback rule, and an
attention budget make different claims and must be reported separately.

TRACE-C retains its historical name, ``temporal relational anomaly detection
with copula calibration,'' but this manuscript narrows that description to the
shipped algorithm. The method performs strictly-prior online scoring with
rolling robust-z marginals and rank calibration. Its relational channel has the
algebraic form of a Gaussian copula log-density contrast \citep{song2000gaussiancopula},
but it is evaluated on robust-z residuals without a probability-integral or
normal-score transform. It is therefore not a literal copula density. We keep
the TRACE-C name for continuity and use ``rank-calibrated relational anomaly
detection'' as the scientifically accurate description.

The output $p$-values have an equally limited interpretation. They are
upper-tail conformal/rank values used to order and select windows. They are not
the probability that an alert is genuine, not posterior confidence, and not a
guarantee of time-series coverage. Exchangeability provides a useful benchmark
for interpreting ranks, but it is not established for these data.

This paper makes five contributions:
\begin{enumerate}
  \item It specifies an auditable three-channel detector that preserves
  residual magnitude while making every reference strictly prior to the
  scored window.
  \item It distinguishes the Gaussian copula-form dependence contrast actually
  implemented from literal copula calibration, and distinguishes the Fisher
  aggregation score from a chi-square test.
  \item It reports the full selection path---ordinary BH attempted first, a
  disclosed record-rule fallback, and a fixed-block alert budget---together
  with the assumptions that prevent a nominal FDR claim.
  \item It reports a frozen 2020 run, a post-hoc non-identical baseline
  comparison, and a 2019 development-year channel ablation showing that
  Atiyah's rank-1 result is local-led rather than copula-only, and that
  Fisher fusion can bury the short frequency event relative to the temporal
  channel alone.
  \item It ships the evaluation as an evidence-bound package: the TRACE-C
  hold-out was frozen before inspection, retained inputs are checksummed, and
  every table and figure in this manuscript is generated from committed
  machine-readable reports rather than transcribed by hand.
\end{enumerate}

The results precede the limitations and conclusion so that the honesty ledger
can be read against concrete evidence. TRACE-C ranks Storm Atiyah first in
2019 development, but the ablation does not support a joint-surprise account
of that window. Reconstruction methods isolate the short 2019 frequency event
better. The frozen 2020 run produces no selected alerts. These outcomes define
the operating envelope supported by the present evidence.

%% file: sections/related-work.tex
\section{Related work}

\subsection{Rank calibration under dependence}

Conformal prediction obtains finite-sample rank guarantees from symmetry or
exchangeability arguments \citep{shafer2008tutorial}. Operational telemetry
generally has seasonality, autocorrelation, and changing regimes, so a correct
arithmetic rank formula does not by itself supply those assumptions. Methods
for dependent data can instead use block permutations or related constructions
under explicit dependence conditions \citep{chernozhukov2018dependentconformal}.
TRACE-C does not implement the block method of
\citet{chernozhukov2018dependentconformal}; the citation identifies a relevant
alternative, not a theorem inherited by this implementation.

Conformal outlier testing connects rank $p$-values to multiple testing and can
obtain FDR guarantees in a setting with independent reference and test samples
and a carefully characterized dependence structure \citep{bates2023conformaloutliers}.
That reference/test setting does not validate TRACE-C's growing, reused online
reference. Here the ranks are transparent selection statistics and the
observed-versus-benchmark counts are empirical diagnostics.

\subsection{Dependence scores and evidence aggregation}

Sklar's theorem separates multivariate dependence from continuous marginals
through probability-integral transforms \citep{sklar1959fonctions}. Gaussian
copula models then express dependence through normal scores and a correlation
matrix \citep{song2000gaussiancopula}. TRACE-C borrows the resulting quadratic
contrast but applies it to robust-z residuals. The absence of the marginal
transform is material: ``copula-form'' describes the algebra, not a fitted
copula density.

Fisher's statistic combines small $p$-values through a sum of log terms
\citep{fisher1932statisticalmethods}. Its familiar chi-square calibration
requires conditions not asserted for TRACE-C's dependent channel ranks. We use
the construction only as an aggregation score and calibrate that score again
by its online rank.

Ordinary Benjamini--Hochberg (BH) controls FDR under independence and has known
extensions under suitable positive dependence conditions
\citep{benjamini1995fdr}. Those independence/PRDS conditions are not
established here. General dependence motivates more conservative procedures
such as Benjamini--Yekutieli \citep{benjamini2001dependency}, but TRACE-C does
not implement that correction. Consequently, ``BH'' below means a nominal
selection attempt, not demonstrated FDR control.

\subsection{Robust scaling, records, and comparator families}

Median and MAD scaling provides resistance to extreme observations
\citep{rousseeuw1993mad}. The fallback rule selects a new upper record; under an
exchangeable continuous sequence its expected number of records follows the
classical harmonic calculation \citep{renyi1962records}. Serial dependence,
ties, and drift prevent that expected count from being a guarantee in the
present application.

The post-hoc comparison represents four common anomaly-score families:
nonlinear reconstruction by an autoencoder
\citep{sakurada2014autoencoders,vijay2020kerasautoencoder}, PCA reconstruction
residuals \citep{jackson1979pcaresiduals}, Isolation Forest
\citep{liu2008isolationforest}, and the Spectral Residual method
\citep{ren2019spectralresidual}. These citations motivate the detector classes;
the exact fitted configurations and online rank envelope are described in
Section~\ref{sec:protocol}.

%% file: sections/method.tex
\section{Method}

\subsection{Problem setup and online information set}

Let $x_{t,j}$ denote aligned telemetry from stream $j\in\{1,\ldots,d\}$ at
half-hour row $t$. TRACE-C partitions the row sequence into non-overlapping
windows $I_w=\{wW,\ldots,wW+W-1\}$ with $W=4$, corresponding to two hours on
ordinary 48-period days. A score for $I_w$ may use fitted parameters from the
designated fit segment and observations with indices strictly smaller than the
row or window being scored. We call this \emph{strictly-prior online scoring};
it is an information-flow property, not a causal-inference claim.

The desired output is a ranked list of anomalous windows together with compact
score contributions. No event annotation enters the scoring function. The
implementation emits only the window index $w$, starting row $t_0$, outer rank
$p$, aggregate score $S$, channel contributions, and per-stream contributions.
It does not implement a separate severity field or a reason-code schema.

\subsection{Regime-conditioned robust residuals}

Each row is assigned a regime given by settlement period crossed with
weekday/weekend status. For every stream and regime, the method takes the
$K=40$ most recent same-regime values strictly before $t$, computes their
median $m_{t,j}$ and median absolute deviation $d_{t,j}$, and forms
\begin{equation}
 z_{t,j}=\operatorname{clip}\!\left(
 \frac{x_{t,j}-m_{t,j}}{1.4826\,d_{t,j}},-10,10\right).
 \label{eq:robust-z}
\end{equation}
MAD scaling follows the standard robust-scale construction
\citep{rousseeuw1993mad}; a zero MAD is replaced in code by a small positive
floor.

This magnitude-preserving choice is deliberate. A same-regime rank-PIT with
$K$ reference values followed by an inverse-normal map cannot exceed
\(\Phi^{-1}((K+\tfrac12)/(K+1))\). At $K=40$ the ceiling is approximately
$2.2509$, so a barely new record and a much deeper physical excursion receive
the same transformed magnitude. Equation~\eqref{eq:robust-z} retains magnitude
up to the disclosed $\pm10$ clipping limit; the later rank layers still govern
selection.

\subsection{Three window channels}

The shipped detector contains exactly three channels. First, the local channel
is the largest absolute normalized window sum,
\begin{equation}
 L_w=\max_j\left|\frac{1}{\sqrt W}\sum_{t\in I_w}z_{t,j}\right|.
 \label{eq:local}
\end{equation}
It favors sustained magnitude within at least one stream while avoiding a sum
over streams that would conceal the responsible sensor.

Second, let $R$ be the normalized residual cross-product matrix estimated from
complete robust-z rows in the January--April fit segment, with a ridge added if
needed for a Cholesky factorization. The relational score averages
\begin{equation}
 G_w=\frac{1}{W}\sum_{t\in I_w}
 \left\{\frac12\bigl(z_t^{\mathsf T}R^{-1}z_t-z_t^{\mathsf T}z_t\bigr)
       +\frac12\log|R|\right\}.
 \label{eq:relation}
\end{equation}
This is a Gaussian copula-form negative log-density contrast suggested by the
usual Gaussian copula algebra \citep{sklar1959fonctions,song2000gaussiancopula}.
However, the inputs are the robust residuals in
Equation~\eqref{eq:robust-z}; the implementation performs no PIT and no
normal-score transform. Thus $G_w$ is not a literal copula density, and its
name is historical rather than a calibration claim.

Third, an AR(1) coefficient $\phi_j$ and innovation standard deviation
$\sigma_{e,j}$ are fitted per stream on the same designated fit segment. The
temporal channel is the worst standardized innovation within the window,
\begin{equation}
 T_w=\max_{j,t\in I_w}
 \frac{|z_{t,j}-\phi_jz_{t-1,j}|}{\sigma_{e,j}}.
 \label{eq:temporal}
\end{equation}
Taking a maximum rather than a mean preserves a sharp transition that would be
diluted by the other rows of a two-hour window.

\subsection{Trailing channel ranks and outer rank}

For each channel $c\in\{L,G,T\}$, let $\mathcal R_{c,w}$ be the trailing 240
eligible channel scores strictly before window $w$. Once the reference is full,
the upper-tail channel value is
\begin{equation}
 p_{c,w}=\frac{1+\#\{r\in\mathcal R_{c,w}:r\geq C_{c,w}\}}
 {|\mathcal R_{c,w}|+1}.
 \label{eq:channel-rank}
\end{equation}
The three channel ranks are dependent because they summarize the same rows.
TRACE-C constructs
\begin{equation}
 S_w=-2\sum_{c\in\{L,G,T\}}\log p_{c,w},
 \label{eq:fisher}
\end{equation}
following Fisher's aggregation form \citep{fisher1932statisticalmethods}.
Equation~\eqref{eq:fisher} is only an aggregation score: no chi-square null
distribution is claimed.

The outer value ranks $S_w$ against the growing sorted set
$\mathcal S_w$ of all strictly-prior available aggregate scores. It is emitted
after at least 40 prior $S$ values:
\begin{equation}
 p_w=\frac{1+\#\{s\in\mathcal S_w:s\geq S_w\}}
 {|\mathcal S_w|+1},\qquad |\mathcal S_w|\geq40.
 \label{eq:outer-rank}
\end{equation}
All comparisons for selection use the exact arithmetic value from
Equation~\eqref{eq:outer-rank}, not a displayed decimal. These are
conformal/rank $p$-values for ordering and selection. Their finite-sample
formula is exact, but the usual distributional interpretation still depends
on exchangeability \citep{shafer2008tutorial}.

\subsection{Selection, fallback, and budget}

Across the scored windows in an evaluation segment, the implementation first
applies ordinary BH at nominal $q=0.05$ \citep{benjamini1995fdr}. Independence
or PRDS for these reused online ranks is not established, so this step is only
a nominal selection attempt. If and only if BH returns zero windows, the code
falls back to the upper-record rule
\begin{equation}
 p_w\leq p^{\min}_w=\frac{1}{|\mathcal S_w|+1},
\end{equation}
which is equivalent to $S_w$ strictly exceeding every earlier aggregate score.
The fallback is commonly activated because rank granularity makes early BH
thresholds unreachable. The report
always names the rule used.

Under an exchangeable continuous score sequence, the record indicator has
expectation $1/(|\mathcal S_w|+1)$ \citep{renyi1962records}. TRACE-C reports
the sum of those expectations as an \emph{exchangeable-continuous benchmark},
not as a guarantee for seasonal, autocorrelated telemetry. Finally, selected
windows are grouped by fixed 48-row blocks and at most two are retained per
block. This is an attention budget, not a true calendar-day or daylight-saving
safe construction.

%% file: sections/data-protocol.tex
\section{Data and evaluation protocol}
\label{sec:protocol}

\subsection{Streams, aggregation, and provenance}

The case study uses public Great Britain electricity-system telemetry from the
National Energy System Operator (NESO). Five demand-side columns are read at
half-hour settlement resolution: national demand (ND), transmission-system
demand (TSD), embedded wind generation, embedded solar generation, and pumped
storage pumping \citep{neso2019historicdemand}. A sixth column is formed from
NESO system-frequency data by taking the maximum absolute deviation from
50~Hz within each settlement period \citep{neso2019systemfrequency}. This
half-hour maximum retains an extreme deviation but not its sub-period timing or
waveform.

The retained 2019 and 2020 inputs are pinned by byte count and SHA-256 checksum.
The report objects identify the generating source files and source-manifest
hashes; the paper tables are generated from those committed reports. The data
are used under the NESO Open Data Licence \citep{nesoOpenLicence}. Required
attribution is: ``Supported by National Energy SO Open Data.''

Rows are ordered by settlement date and period. The same-regime residual
history distinguishes weekday from weekend and settlement period. The loader
handles settlement-period irregularities defensively, while the later alert
budget still groups fixed 48-row blocks; that distinction matters on
daylight-saving transitions.

\subsection{Segments and frozen protocol}

January--April 2019 is designated as the fit segment for the dependence matrix
and AR(1) parameters. It is assumed representative enough for this role; it is
not established to be anomaly-free or ``clean.'' May--June supplies prior
windows before the declared evaluation segment. July--December 2019 is the
development segment, comprising 2,208 scored non-overlapping windows. Choices
of $W=4$, $K=40$, the robust-z marginal, and inclusion of the frequency summary
were informed by 2019 evidence and are not validation outcomes.

The method, sensor set, and configuration were then fixed before the 2020
events were inspected. Rolling references continue across the year boundary,
and all 4,392 scored windows in 2020 form the frozen TRACE-C hold-out. We use
``frozen'' to describe this development chronology, not to assert that the
hold-out is a clean null year. In particular, 2020 contains major weather and
demand-regime changes documented independently
\citep{metoffice2020stormseason,ukgov2020lockdown,ngeso2021endyear}.

\subsection{Event annotations and opportunities}

Annotations are joined only after ranking. The 9 August 2019 power cut and
frequency disturbance is documented by Ofgem \citep{ofgem2020poweroutage}; its
evaluation interval overlaps 5 two-hour windows. Storm Atiyah spans 12 windows,
while the two-day Ciara and Dennis intervals contain 24 windows each, using the
Met Office storm chronology \citep{metoffice2020stormseason}. The first
lockdown evaluation interval is 23 March--5 April 2020, or 168 windows, anchored
to the 23 March announcement \citep{ukgov2020lockdown}. Because event intervals
differ, their best ranks have unequal opportunity for a small value and should
not be treated as like-for-like point-event metrics.

Storm Ellen and Storm Alex were not pre-specified evaluation intervals. Their
names were attached after ranking to the 20 August and 3 October windows using
the external storm record \citep{metoffice2020stormseason,metoffice2020stormalex}.
They are interpretations of unalerted windows, not detector inputs or
discoveries.

\subsection{Post-hoc baselines}

Four baselines share the source rows, six-stream sensor set, $W=4$ windows,
fit cutoff, evaluation dates, score direction, and event intervals. They are a
convolutional autoencoder inspired by standard reconstruction approaches
\citep{sakurada2014autoencoders,vijay2020kerasautoencoder}, PCA reconstruction
with three components \citep{jackson1979pcaresiduals}, a 200-tree Isolation
Forest with fixed seed on window mean/standard-deviation features
\citep{liu2008isolationforest}, and a trailing-context Spectral Residual score
\citep{ren2019spectralresidual}.

Their protocol is explicitly non-identical to TRACE-C. Each baseline uses
global per-stream fit mean/standard-deviation scaling, then directly ranks its
raw anomaly score against a growing strictly-prior history after a 40-window
post-fit warm-up. Selection is an unbudgeted record rule. TRACE-C instead uses
same-regime robust residuals, three trailing 240-window channel ranks, Fisher
aggregation, a growing outer rank, a BH-first selector, and a two-per-block
budget. The baseline suite was constructed after TRACE-C's 2020 results were
known. Thus only TRACE-C was frozen before 2020 inspection, and Table
\ref{tab:baselines} is a post-hoc comparison rather than a second hold-out
experiment.

\subsection{Reproducibility}

The deterministic TypeScript evaluation rebuilds both TRACE-C reports from the
pinned inputs. The baseline score artifact records its fixed random seed,
trainer/source hash, input-series hash, runtime versions, and training history;
the final evaluator verifies those fields before producing the comparison.
Generated \LaTeX{} tables are derived from the JSON evidence rather than edited
manually. These measures support computational reproduction of the reported
arithmetic, while making no claim that different platforms will produce
bitwise-identical external-model artifacts.

%% file: sections/results.tex
\section{Results}
\label{sec:results}

\subsection{Empirical rank diagnostics and selection}

Table~\ref{tab:trace-calibration} compares the number of outer rank values below
selected thresholds with the arithmetic count $\alpha n$. In 2019, the full
report also gives 51 observed versus 44.2 at $p\leq.02$. The displayed
thresholds are close in some segments and conservative in others: 2019 has
120/110.4 at $.05$ and 23/22.1 at $.01$; pre-COVID 2020 has 50/45.0 and 3/9.0.
The full 2020 count is 205/219.6 at $.05$ (and 39/43.9 at $.01$ in the
machine-readable report). These are descriptive empirical rank diagnostics.
Autocorrelation, drift, reused references, and known events rule out treating
them as proof of uniform $p$-values, coverage, or false-discovery control
\citep{shafer2008tutorial,bates2023conformaloutliers}.
Figure~\ref{fig:rank-diagnostics} shows the same diagnostics across all five
detectors under the shared scoring envelope.

\input{generated/fig-rank-diagnostics}

BH selected no windows in either reported TRACE-C segment, so the disclosed
record fallback ran. The 2019 development segment contains two selected alerts
versus an exchangeable-continuous benchmark of 1.19; one is the Atiyah window
and the other is an unlabelled 21 July window. The frozen 2020 segment contains
zero selected alerts versus a benchmark of 0.87. Neither comparison is a
hypothesis-test guarantee \citep{renyi1962records}. The absence of later records
despite several small rank values is evidence of record saturation: once a high
aggregate score enters the growing history, the rule becomes increasingly hard
to trigger. Zero alerts is therefore a substantive frozen result, not evidence
that 2020 lacked unusual windows. Figure~\ref{fig:holdout-timeline} plots the
full frozen segment with the post-hoc event annotations.

\input{generated/fig-holdout-timeline}

\begingroup
\AtBeginEnvironment{tabular}{%
  \ifnum\value{table}=3
    \tiny\setlength{\tabcolsep}{2pt}%
  \else
    \scriptsize\setlength{\tabcolsep}{1pt}%
  \fi
}
\input{generated/results-table}
\endgroup

\subsection{2019 annotated event ranks}

The 2019 results are sharply mixed. Storm Atiyah is rank 1 of 2,208 with
$p=0.000350$ and passes the record rule. The full detector's lead channel on
that window is local, not copula-form, and the episode is a sustained
multi-stream extreme documented by the Met Office
\citep{metoffice2020stormseason}. By contrast, the 9 August power-cut interval
has a best rank of 143 and $p=0.060420$ across 5 opportunities, despite the
frequency-deviation stream. The source event was real and operationally
important \citep{ofgem2020poweroutage}; importance and separability at the
chosen aggregation are different properties.

\subsection{2019 channel ablation}

\input{generated/ablation-table}

Table~\ref{tab:ablation} is a development-year diagnostic, not a hold-out: the
full detector's Atiyah lead channel was inspected before the reruns. Isolating
the copula-form channel does not recover Atiyah (rank 59). Dropping that
channel leaves Atiyah at rank 2 without a record-rule alert, so the copula term
changes whether the fallback fires, not whether the window is extreme. Local
is the Atiyah engine (local-only rank 3; dropping local removes alerts). The
short frequency event is strongest on the temporal channel alone (rank 40
versus 143 fused). Fisher aggregation can therefore bury a brief transient
that a single channel would rank more highly. These reruns do not support
describing Atiyah as a copula-only discovery.

\subsection{Frozen 2020 hold-out event ranks}

In the frozen 2020 run, Ciara's best of 24 windows is rank 44 with
$p=0.012058$, and Dennis's best of 24 is rank 137 with $p=0.033217$. These storm
intervals follow the external chronology \citep{metoffice2020stormseason}. The
lockdown annotation covers 168 opportunities from 23 March through 5 April
\citep{ukgov2020lockdown}. TRACE-C's best lockdown-window result is rank 45,
$p=0.012559$, at 06:00 on 28 March. It is not the onset window and must not be
described as onset detection.

The highest-ranked 2020 window, 20 August at 03:00, was later interpreted as
Storm Ellen; it has $p=0.000504$ but is unalerted. Rank 2, 20 January at 14:00,
is unlabelled. Rank 5, 3 October at 23:00, was later interpreted as Storm Alex
using the external storm account \citep{metoffice2020stormalex}; it too is
unalerted. Table~\ref{tab:post-ranked-weather} lists the two post-ranking weather
interpretations without implying that they are the top two windows or
algorithmic discoveries.

\subsection{Post-hoc baseline comparison}

\input{generated/baseline-table}

Table~\ref{tab:baselines} reinforces the aggregation trade-off. The
convolutional autoencoder and PCA reconstruction score the 2019 frequency event
at rank 1, while Isolation Forest and Spectral Residual give rank 2. These
detector families target reconstruction, isolation, or spectral saliency
rather than TRACE-C's relational rank aggregate
\citep{sakurada2014autoencoders,jackson1979pcaresiduals,liu2008isolationforest,ren2019spectralresidual}.
TRACE-C instead gives Atiyah rank 1; the autoencoder, PCA, Isolation Forest, and
Spectral Residual give Atiyah ranks 34, 12, 3, and 26, respectively.

For the 168-window lockdown interval, Isolation Forest ranks its best window
first, PCA 22nd, the autoencoder 46th, and Spectral Residual 6th, compared with
TRACE-C at 45th. The best baseline dates are 5 April for the autoencoder, PCA,
and Isolation Forest, and 28 March for Spectral Residual. These are interval
search outcomes, not claims about 23 March onset. The pre-COVID $.05$ counts
also vary widely: 27/45 for the autoencoder, 82/45 for PCA, 49/45 for Isolation
Forest, and 93/45 for Spectral Residual. Because the baseline ranks use a global
regime-naive reference and were produced after inspection of TRACE-C's 2020
results, the table is neither a matched selection-budget leaderboard nor an
independent validation.

Overall, no detector dominates. TRACE-C's fused detector ranks Atiyah first,
but the ablation shows that rank is not copula detection. Reconstruction is
far better on the short frequency event, consistent with the temporal-only
ablation ranking that event 40th while the fused detector ranks it 143rd.
Isolation Forest supplies the smallest lockdown-interval rank. The appropriate
interpretation is an interaction among score construction, interval length,
sensor set, and temporal resolution, not a universal ranking of algorithms.

%% file: generated/fig-rank-diagnostics.tex
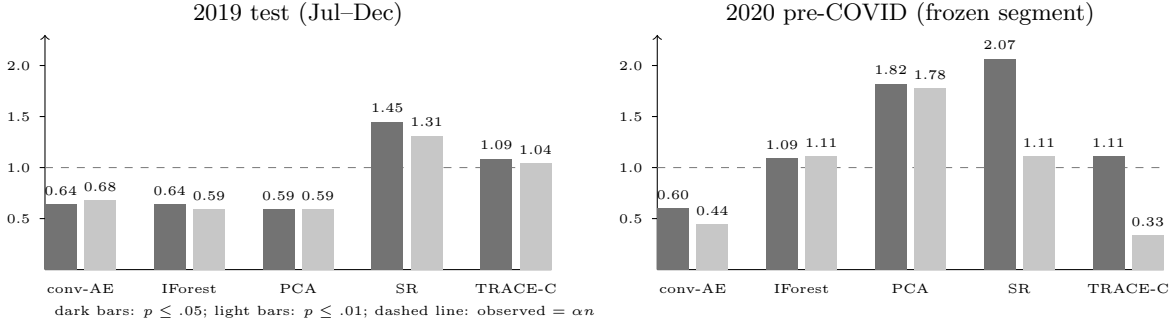
\begin{figure*}[t]
\centering
\begin{tikzpicture}
\draw[->] (0,0) -- (0,3.105);
\draw (0,0) -- (6.700,0);
\draw (0,0.675) -- (-0.07,0.675) node[left,font=\tiny] {0.5};
\draw (0,1.350) -- (-0.07,1.350) node[left,font=\tiny] {1.0};
\draw (0,2.025) -- (-0.07,2.025) node[left,font=\tiny] {1.5};
\draw (0,2.700) -- (-0.07,2.700) node[left,font=\tiny] {2.0};
\draw[dashed,gray] (0,1.350) -- (6.700,1.350);
\fill[black!55] (0.000,0) rectangle (0.420,0.868);
\node[above,font=\tiny] at (0.210,0.868) {0.64};
\fill[black!22] (0.520,0) rectangle (0.940,0.916);
\node[above,font=\tiny] at (0.730,0.916) {0.68};
\node[below,font=\tiny] at (0.470,-0.06) {conv-AE};
\fill[black!55] (1.440,0) rectangle (1.860,0.868);
\node[above,font=\tiny] at (1.650,0.868) {0.64};
\fill[black!22] (1.960,0) rectangle (2.380,0.794);
\node[above,font=\tiny] at (2.170,0.794) {0.59};
\node[below,font=\tiny] at (1.910,-0.06) {IForest};
\fill[black!55] (2.880,0) rectangle (3.300,0.795);
\node[above,font=\tiny] at (3.090,0.795) {0.59};
\fill[black!22] (3.400,0) rectangle (3.820,0.794);
\node[above,font=\tiny] at (3.610,0.794) {0.59};
\node[below,font=\tiny] at (3.350,-0.06) {PCA};
\fill[black!55] (4.320,0) rectangle (4.740,1.957);
\node[above,font=\tiny] at (4.530,1.957) {1.45};
\fill[black!22] (4.840,0) rectangle (5.260,1.771);
\node[above,font=\tiny] at (5.050,1.771) {1.31};
\node[below,font=\tiny] at (4.790,-0.06) {SR};
\fill[black!55] (5.760,0) rectangle (6.180,1.467);
\node[above,font=\tiny] at (5.970,1.467) {1.09};
\fill[black!22] (6.280,0) rectangle (6.700,1.405);
\node[above,font=\tiny] at (6.490,1.405) {1.04};
\node[below,font=\tiny] at (6.230,-0.06) {TRACE-C};
\node[above,font=\small] at (3.350,3.105) {2019 test (Jul--Dec)};
\begin{scope}[xshift=8.100cm]
\draw[->] (0,0) -- (0,3.105);
\draw (0,0) -- (6.700,0);
\draw (0,0.675) -- (-0.07,0.675) node[left,font=\tiny] {0.5};
\draw (0,1.350) -- (-0.07,1.350) node[left,font=\tiny] {1.0};
\draw (0,2.025) -- (-0.07,2.025) node[left,font=\tiny] {1.5};
\draw (0,2.700) -- (-0.07,2.700) node[left,font=\tiny] {2.0};
\draw[dashed,gray] (0,1.350) -- (6.700,1.350);
\fill[black!55] (0.000,0) rectangle (0.420,0.810);
\node[above,font=\tiny] at (0.210,0.810) {0.60};
\fill[black!22] (0.520,0) rectangle (0.940,0.600);
\node[above,font=\tiny] at (0.730,0.600) {0.44};
\node[below,font=\tiny] at (0.470,-0.06) {conv-AE};
\fill[black!55] (1.440,0) rectangle (1.860,1.470);
\node[above,font=\tiny] at (1.650,1.470) {1.09};
\fill[black!22] (1.960,0) rectangle (2.380,1.500);
\node[above,font=\tiny] at (2.170,1.500) {1.11};
\node[below,font=\tiny] at (1.910,-0.06) {IForest};
\fill[black!55] (2.880,0) rectangle (3.300,2.460);
\node[above,font=\tiny] at (3.090,2.460) {1.82};
\fill[black!22] (3.400,0) rectangle (3.820,2.400);
\node[above,font=\tiny] at (3.610,2.400) {1.78};
\node[below,font=\tiny] at (3.350,-0.06) {PCA};
\fill[black!55] (4.320,0) rectangle (4.740,2.790);
\node[above,font=\tiny] at (4.530,2.790) {2.07};
\fill[black!22] (4.840,0) rectangle (5.260,1.500);
\node[above,font=\tiny] at (5.050,1.500) {1.11};
\node[below,font=\tiny] at (4.790,-0.06) {SR};
\fill[black!55] (5.760,0) rectangle (6.180,1.500);
\node[above,font=\tiny] at (5.970,1.500) {1.11};
\fill[black!22] (6.280,0) rectangle (6.700,0.450);
\node[above,font=\tiny] at (6.490,0.450) {0.33};
\node[below,font=\tiny] at (6.230,-0.06) {TRACE-C};
\node[above,font=\small] at (3.350,3.105) {2020 pre-COVID (frozen segment)};
\end{scope}
\node[font=\tiny,anchor=west] at (0,-0.55)
  {dark bars: $p\leq.05$; light bars: $p\leq.01$; dashed line: observed $=\alpha n$};
\end{tikzpicture}
\caption{Empirical rank diagnostics under the shared scoring envelope:
observed counts of outer rank values at thresholds $.05$ and $.01$, expressed
as a ratio to the arithmetic benchmark $\alpha n$ (dashed line). Values are
read directly from the committed reports. Descriptive only: autocorrelation,
drift, reused references, and known events preclude reading these as uniform
$p$-value or false-discovery guarantees (Section~\ref{sec:results}).}
\label{fig:rank-diagnostics}
\end{figure*}

%% file: generated/results-table.tex
\begin{table}[t]
\centering
\caption{Empirical rank-p diagnostics. Expected counts are exchangeable-score references, not time-series coverage guarantees.}
\label{tab:trace-calibration}
\begin{tabular}{@{}lr@{}}
\toprule
Segment and threshold & Observed/reference \\
\midrule
2019 development, $p\leq .05$ & 120/110.4 \\
2019 development, $p\leq .01$ & 23/22.1 \\
2020 pre-COVID, $p\leq .05$ & 50/45.0 \\
2020 pre-COVID, $p\leq .01$ & 3/9.0 \\
2020 full year, $p\leq .05$ & 205/219.6 \\
\bottomrule
\end{tabular}
\end{table}

\begin{table*}[t]
\centering
\caption{TRACE-C event-window ranks. Labels are used only for post-hoc evaluation.}
\label{tab:trace-events}
\begin{tabular}{@{}llrrr@{}}
\toprule
Annotation & Segment & Best rank & Best $p$ & Opportunity windows \\
\midrule
GB frequency event & 2019 development & 143 & 0.060420 & 5 \\
Storm Atiyah & 2019 development & 1 & 0.000350 & 12 \\
Storm Ciara & 2020 frozen hold-out & 44 & 0.012058 & 24 \\
Storm Dennis & 2020 frozen hold-out & 137 & 0.033217 & 24 \\
Lockdown transition (14 days; best 2020-03-28 06:00) & 2020 frozen hold-out & 45 & 0.012559 & 168 \\
\bottomrule
\end{tabular}
\end{table*}

\begin{table}[t]
\centering
\caption{Externally interpreted, top-ranked 2020 windows. Neither name was a detector annotation and no 2020 window passed the frozen record rule.}
\label{tab:post-ranked-weather}
\begin{tabular}{@{}lrrl@{}}
\toprule
External interpretation & Rank & Timestamp & $p$ \\
\midrule
Storm Ellen (post-ranking interpretation) & 1 & 2020-08-20 03:00 & 0.000504 \\
Storm Alex (post-ranking interpretation) & 5 & 2020-10-03 23:00 & 0.001539 \\
\bottomrule
\end{tabular}
\end{table}

%% file: generated/ablation-table.tex
\begin{table*}[t]
\centering
\caption{2019 development-year channel ablation, not a hold-out. The full detector's Atiyah lead channel was inspected before these reruns. Lower rank is better.}
\label{tab:ablation}
\begin{tabular}{@{}llrrr@{}}
\toprule
Variant & Channels & Atiyah rank & Blackout rank & Alerts \\
\midrule
full (L+G+T) & local+copula+temporal & 1 (alert) & 143 & 2 \\
drop copula & local+temporal & 2 & 69 & 1 \\
drop local & copula+temporal & 3 & 119 & 0 \\
drop temporal & local+copula & 5 & 338 & 1 \\
local only & local & 3 & 241 & 0 \\
copula only & copula & 59 & 323 & 0 \\
temporal only & temporal & 10 & 40 & 0 \\
\bottomrule
\end{tabular}
\end{table*}

%% file: generated/baseline-table.tex
\begin{table*}[t]
\centering
\caption{Post-hoc detector comparison. Lower event rank is better. The lockdown column searches a 168-window interval; parenthesized dates show each detector's best window. Protocols are not identical and only TRACE-C was frozen before 2020 inspection.}
\label{tab:baselines}
\begin{tabular}{@{}lrrrrrl@{}}
\toprule
Detector & Blackout & Atiyah & Ciara & Dennis & Lockdown window & Pre-COVID $p\leq .05$ \\
\midrule
TRACE-C & 143 & 1 & 44 & 137 & 45 (2020-03-28) & 50/45.0 \\
Convolutional autoencoder & 1 & 34 & 1 & 29 & 46 (2020-04-05) & 27/45.0 \\
PCA reconstruction & 1 & 12 & 12 & 34 & 22 (2020-04-05) & 82/45.0 \\
Isolation Forest & 2 & 3 & 10 & 178 & 1 (2020-04-05) & 49/45.0 \\
Spectral Residual & 2 & 26 & 12 & 38 & 6 (2020-03-28) & 93/45.0 \\
\bottomrule
\end{tabular}
\end{table*}

%% file: sections/limitations.tex
\section{Limitations and honesty ledger}

The central limitation is calibration under dependence. The rank arithmetic is
strictly prior, but grid telemetry is not shown to be exchangeable. The rolling
240-window channel references adapt to local score distributions; they do not
turn the sequence into independent or identically distributed observations.
The growing outer reference also reuses history across many tests. Consequently
the empirical rank counts are descriptive, ordinary BH is nominal, and neither
coverage nor FDR is established. Block conformal methods provide one relevant
research direction \citep{chernozhukov2018dependentconformal}, but no such block
construction is implemented here. Likewise, the conditions studied for
conformal outlier testing do not transfer automatically to this online reuse
pattern \citep{bates2023conformaloutliers}.

The record fallback is transparent but weak on a long horizon. Its expected
count is an exchangeable-continuous benchmark, not a time-series guarantee
\citep{renyi1962records}, and an early extreme can suppress later records. The
zero-alert 2020 result alongside low-ranked unalerted windows empirically
exhibits this saturation. A future selector should be specified before a new
untouched period and evaluated with dependence and finite-resolution effects in
view; this paper does not retroactively substitute such a rule.

Development history also limits inference. A historical fixed-calibration
episode produced 62 alerts; about 90\% were judged seasonal-drift artifacts,
corresponding to roughly 20-fold anti-conservatism. This is a narrative
disclosure only: there is no shipped artifact for that old run, and it is not a
current comparator. The committed current comparison contains the demand-only
variant, not a reconstruction of the historical fixed-calibration episode.
Within the present algorithm, $W\in\{2,4\}$ and $K\in\{20,40\}$ were considered
using 2019 development evidence, the change from rank-PIT to robust-z was
event-informed, and the frequency stream was added after demand-only analysis.
Freezing 2020 limits subsequent adaptation but cannot erase those choices.

The relational terminology needs similar restraint. Equation
\ref{eq:relation} resembles a Gaussian copula log-density contrast, yet the
robust-z inputs are not probability-integral transformed normal scores as in a
literal copula construction \citep{sklar1959fonctions,song2000gaussiancopula}.
The score may still be useful as a fitted dependence contrast, but copula
likelihood theory does not directly calibrate it. The three channel ranks are
also dependent, so Fisher's chi-square reference is deliberately not used
\citep{fisher1932statisticalmethods}.

Sensor selection and aggregation define detectability. The frequency stream
keeps only the maximum absolute deviation within a half-hour, and all channels
are evaluated in non-overlapping two-hour windows. A short transient can lose
its timing and be dominated by sustained background extremes, whereas a broad
weather episode can produce coherent contributions across several streams.
The stream-contribution list discloses large window residuals but is descriptive
rather than a mechanistic explanation. A 2019 development-year channel ablation
is reported; it is peek-informed and is not a second hold-out. Missing-stream
robustness, alternative aggregations, synthetics, PR-AUC, and matched-budget
comparisons have not been evaluated.

The baseline suite is post-hoc and protocol-non-identical. Global scaling,
direct growing raw-score ranks, and unbudgeted record selection differ from
TRACE-C's regime conditioning, two-stage ranks, aggregation, BH attempt, and
fixed-block budget. Only TRACE-C was frozen before 2020 inspection, so apparent
wins in either direction are exploratory. Event-best ranks also favor longer
intervals through more opportunities.

The attention limit is two selected windows per fixed 48-row block. It is not a
calendar service: days with 46 or 50 settlement periods and daylight-saving
transitions are not handled as true civil days. A deployment would need a
calendar-aware budget and explicit policy for late or revised telemetry.

Finally, the governance scope is deliberately narrow. TRACE-C ranks process
and sensor anomalies in operational telemetry for human review. It is not
designed, evaluated, or authorized for person-risk scoring, and the evidence in
this study cannot support decisions about individuals.

%% file: sections/conclusion.tex
\section{Conclusion}

TRACE-C is best understood as a strictly-prior rank-calibrated detector, not a
literal copula calibration procedure. It combines magnitude-preserving robust
residuals, local/dependence/temporal channels, a rank-aggregated score, and an
explicit selection path. Its rank $p$-values support ordering and selection;
they are not estimated event probabilities or posterior confidence.

The evidence is mixed and operationally informative. TRACE-C ranks Storm
Atiyah first in 2019 development, but the channel ablation shows that this is
a local-led window: copula-only ranks it 59th, and dropping the copula channel
leaves it 2nd without a record alert. Simple reconstruction, and the temporal
channel alone, rank the short frequency event much better than the fused
detector. The frozen 2020 run selects no alerts, while its ranked list contains
plausible weather-associated and unlabelled extremes. The result exposes record
saturation, fusion trade-offs, and sensor/aggregation limits rather than
establishing nominal FDR or copula discovery.

Future work should pre-register a dependence-aware selector, evaluate it on a
new untouched period, preserve sub-period frequency information, test missing
streams and alternative aggregations, and replace fixed 48-row attention blocks
with calendar-aware budgeting. The current contribution is the narrower one:
an auditable algorithm and evidence package whose assumptions, fallbacks, and
negative results are stated at the same resolution as its successes.